\PassOptionsToPackage{table}{xcolor}
\documentclass{article}
\usepackage[preprint]{acl}
\usepackage{times}

\usepackage{amsmath,amsfonts,bm}

\def\eqref#1{equation~\ref{#1}}

\def\1{\bm{1}}

\DeclareMathAlphabet{\mathsfit}{\encodingdefault}{\sfdefault}{m}{sl}
\SetMathAlphabet{\mathsfit}{bold}{\encodingdefault}{\sfdefault}{bx}{n}

\newcommand{\deepscaler}{\textsc{DeepScaleR-1.5B}}

\newcommand{\qwenthreefourb}{\textsc{Qwen3-4B}}
\newcommand{\qwenthreefourbig}{\textsc{Qwen3-4B}}
\newcommand{\qwenthreezeropointsixb}{\textsc{Qwen3-0.6B}}
\newcommand{\qwenthreeosixb}{\textsc{Qwen3-0.6B}}

\usepackage{hyperref}
\usepackage{xurl}
\usepackage{url}
\usepackage{booktabs}       
\usepackage{xcolor}  
\usepackage{array}
\usepackage{multirow}
\usepackage{amsfonts}       
\usepackage{nicefrac}       
\usepackage{microtype}      
\usepackage{graphicx}       
\usepackage{adjustbox}      
\usepackage{float}          
\usepackage{amsmath}        
\usepackage{algorithm}      
\usepackage{algorithmic}    
\usepackage{subcaption}  
\usepackage{dsfont}
\usepackage[utf8]{inputenc}
\usepackage[T1]{fontenc}
\usepackage{listings}
\usepackage{tikz}        
\usetikzlibrary{arrows.meta,positioning,fit,calc,bending}
\usepackage{verbatim}
\usepackage[most,breakable,listings,theorems]{tcolorbox}
\usepackage{dsfont}

\graphicspath{{figures/}{../figures/}}

\title{Real-Time Progress Prediction in Reasoning Language Models} 

\author{Hans Peter Lyngsøe Raaschou-Jensen \and Constanza Fierro \and Anders Søgaard \\
        Department of Computer Science, University of Copenhagen
}

\newcommand{\bestcell}[1]{\textbf{#1}}

\begin{document}
\maketitle
\pagestyle{plain}

\begin{abstract}
Recent reasoning language models, particularly those that employ long latent chains of thought, achieve strong performance on complex agentic tasks. However, as these models operate over increasingly long time horizons, their internal progress becomes opaque to users, making expectation management and real-time oversight difficult. In this work, we investigate whether real-time progress prediction is feasible for such models. We first test whether hidden states encode progress information by discretizing reasoning trajectories and training a linear probe to classify reasoning states. We then fine-tune models to generate progress estimates from 0--100\% during chain-of-thought reasoning. Our strongest progress-reporting checkpoint reaches 0.161 MAE on mathematical reasoning traces and outperforms position baselines in this setting. Finally, we quantify the intrinsic ambiguity of progress labels by measuring how much the implied progress value varies from the same partial rollout. This ambiguity is lowest for \qwenthreefourb{}, whose continuations produce the smallest rollout dispersion, suggesting that larger models can make progress labels more stable by reducing variation in remaining solution length.

\end{abstract}

\section{Introduction}
\label{sec:intro}

Test-time scaling has pushed horizons in natural language processing, leading to reasoning language models, or reasoning LMs, that produce intermediate reasoning traces before answering \citep{deepseekai2025deepseekr1incentivizingreasoningcapability, openai2024openaio1card}. These models introduce substantial user-interaction challenges: the resources consumed by a query can fluctuate significantly, while the model's reasoning progress often remains opaque or unintelligible. This makes it hard for users to know how long they should wait for a task to complete. The problem becomes more salient as language-model agents are evaluated on longer-horizon tasks \citep{kwa2025measuringaiabilitycomplete}. While some work trains models to follow fixed predetermined reasoning budgets \citep{muennighoff2025s1simpletesttimescaling, aggarwal2025l1controllinglongreasoning, intellect22025}, this strategy can fail when the required budget cannot be inferred from the question alone, potentially causing models to return incorrect answers despite being capable of solving the problem given a larger reasoning budget.

Recent research suggests that reasoning LMs encode confidence in their hidden states before materializing a final answer. For example, \citet{zhang2025reasoningmodelsknowtheyre} show that model hidden states contain ``look-ahead'' information, making it possible to predict eventual chain-of-thought correctness before the final answer has been generated, raising the question: Beyond knowing if the solution is reachable, do models also encode the remaining computational distance to the final answer?

Motivated by these results, we test whether reasoning LMs' hidden states contain usable information correlated with partial progress toward the final answer by training a linear probe to predict progress states from the current token. We find that normalized trace position is partially linearly recoverable from hidden states, while noting that this can reflect token-position and trace-length regularities rather than semantic progress alone.

\begin{figure}[t]
\centering
\includegraphics[width=\linewidth]{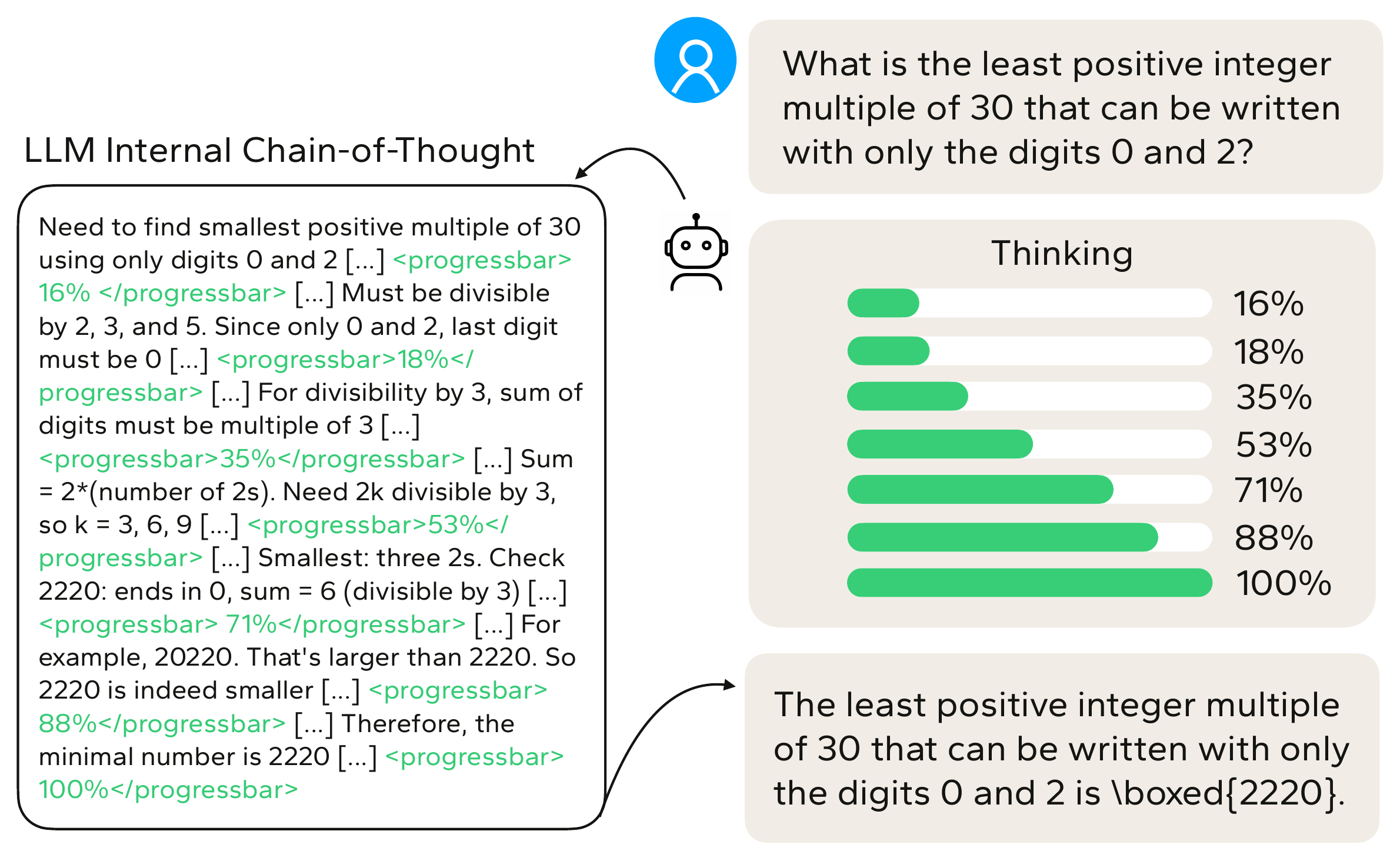}
\caption{Real-time progress tracking during LLM reasoning. The model intermittently updates the user-facing progressbar by using \texttt{<progressbar>..</progressbar>}.}
\label{fig:progress_tracking}
\end{figure}

In summary, our contributions are: (1) We introduce real-time progress prediction for reasoning LMs, enabling real-time visibility into long reasoning trajectories in reasoning language models; (2) We show that reasoning models encode progress information in their internal representations by training linear probes to predict reasoning progress from hidden states; (3) We demonstrate that progress reporting can be incorporated directly during generation; and (4) We analyze how well progress can be measured in stochastic reasoning traces by sampling multiple continuations from shared prefixes, quantifying the range of valid progress values induced by different possible completion lengths.

\section{Related Work}

\paragraph{Search Progress} Beside the foundational work of \citet{thayer2012arewethereyet}, \citet{sudry2021learning} are the first to apply deep learning methods to progress prediction. They convert the problem into a supervised learning task, training an LSTM to predict progress, conditioned on prior search history.  

Beyond general search algorithms, progress estimation has been applied to specific domains. In neural theorem proving, \citet{huang2025leanprogressguidingsearchneural} trained a language model to predict the number of steps remaining to reach proof completion in LEAN \citep{lean4}. Similarly, \citet{ma2019selfmonitoringnavigationagentauxiliary,zhu2020visionlanguagenavigationselfsupervisedauxiliary} have explored progress estimation in vision-language navigation for robotics. 

\paragraph{Internal Reasoning Representations} Recent research suggests that reasoning LMs encode latent beliefs about their reasoning trajectory that are not fully exposed in the generated chain of thought. For example, \citet{zhang2025reasoningmodelsknowtheyre} show that hidden states contain ``look-ahead'' information about whether a chain of thought will eventually be correct before the final answer is generated. Related work finds that representations can predict whether a chain-of-thought process will succeed before completion \citep{afzal2025knowingbeforesaying}, that agreement among latent predictions from intermediate layers distinguishes correct from incorrect reasoning paths \citep{xie2024calibratingreasoninginternalconsistency}, and that final-answer beliefs can become decodable from activations before they are verbalized in the chain of thought \citep{boppana2026reasoningtheater,cox2026decodinganswers}. 

\section{Problem Setup}
\label{definition} 
\paragraph{Reasoning language models} The output of a reasoning LM contains two parts, a reasoning trace and a response. The reasoning trace or Chain-of-Thought (COT) is delimited by specific tokens like \texttt{<think>} and \texttt{</think>}, and contains a sequence of intermediate steps for solving the problem. In our experiments, we use \deepscaler{} \citep{deepscaler2025}, domain-specific reasoning model; and \qwenthreeosixb{} and \qwenthreefourbig{} \citep{qwen3}, general-purpose reasoning models.

\paragraph{Progress Prediction} To our knowledge, progress prediction was systematically studied for the first time in the field of heuristic search algorithms, with \citet{thayer2012arewethereyet} as the seminal work, which defines search progress as follows. Given a search algorithm \(A\) solving a problem \(P\), let \(E(P)\) denote the total number of nodes expanded during search, and let \(\mathrm{Gen}(P)\) denote the number of nodes expanded so far. Search progress is then defined as:
\begin{align}
\mathrm{Prog}(\mathrm{Gen}(P))
&=
\frac{\mathrm{Gen}(P)}
{E(P)} .
\end{align}

\noindent Since \(\mathrm{Gen}(P) \leq E(P)\), progress is naturally normalized to the interval \([0,1]\).

We adapt this formulation to reasoning language models. Given a question \(q\), let
\(
\mathbf{C}_k = (x_1, x_2, \dots, x_k)
\)
denote a partial reasoning trace of length \(k\), where
\(
x_i \sim P_\theta(\cdot \mid q, \texttt{<think>}, x_{<i})
\), and let \(\tau \sim P_\theta(\cdot \mid q, \mathbf{C}_k)\) denote a sampled continuation from that prefix. Here, the observed computation corresponds to the current prefix length \(k\), while the remaining computation corresponds to the expected continuation length conditioned on \(\mathbf{C}_k\). Since exact computation of this expectation is intractable, we approximate it using sampled continuations, yielding:
\[
\mathrm{Progress}(\mathbf{C}_k)
=
\frac{k}
{k + \mathbb{E}_{\tau \sim P_\theta(\cdot \mid q, \mathbf{C}_k)}[|\tau|]}
\approx
\frac{k}{k + |\tau|}.
\]

When comparing possible continuations from the same prefix, we write
\[
\tau^i_\ell \sim P_\theta(\cdot \mid q_i, \mathbf{C}^i_{k})
\]
for rollout continuation \(\ell \in \{1,\dots,r\}\) from prefix \(\mathbf{C}^i_{k}\), where \(r\) is the number of rollouts from a given prefix. The realized progress induced by that continuation is
\[
g_{i,\ell} =
\frac{|\mathbf{C}^i_{k}|}
{|\mathbf{C}^i_{k}| + |\tau^i_\ell|},
\qquad
\bar{g}_i = \frac{1}{r}\sum_{\ell=1}^{r} g_{i,\ell}.
\]

\section{Is Progress Encoded in Hidden States?}\label{sec:probes}

We first investigate whether hidden states in existing reasoning models encode information about reasoning progress. We evaluate this by training linear probes to predict progress from intermediate hidden representations along reasoning trajectories.

\paragraph{Dataset} We construct reasoning datasets by randomly sampling approximately 25K questions without replacement from OpenR1-Math \citep{openr1math2025}. For each model, we obtain one reasoning trace per question using temperature = 0.6 and top\_p = 0.95, following \citep{deepscaler2025}. The public trace datasets use an approximately 90/10 train/test split. For probe training, we reserve all traces longer than 16K tokens from both splits as a held-out dataset used to evaluate length generalization; the remaining traces ($\leq$16K tokens) are used as the in-domain train and evaluation datasets. Split sizes for each model are reported in Appendix~\ref{app:dataset-splits}.

We discretize reasoning progress into 10 labels corresponding to equal bins of normalized trajectory position (0-10\%, 11-20\%, ..., 91-100\%). Formally, for a reasoning trace of length $m$, each token at position $k$ is assigned a label by quantizing the normalized position $k/m$ into one of 10 intervals:
$Q_q = [\frac{q-1}{10}, \frac{q}{10})$ for $q = 1, \ldots, 10$.
A token at position $k$ receives label $q$ if $\frac{k}{m} \in Q_q$.

\paragraph{Probes}
We train two linear probes over hidden representations. Let \(h_{i,j} \in \mathbb{R}^d\) denote the hidden state at token position \(j\) in reasoning trace \(i\).

The first probe, \(f_{\text{token}} : \mathbb{R}^d \rightarrow \mathbb{R}^Q\), predicts the progress bucket directly from the current hidden state \(h_{i,j}\). The second probe, \(f_{\text{q+token}} : \mathbb{R}^{(n+1)d} \rightarrow \mathbb{R}^Q\), predicts the progress bucket from the concatenation \([h_{i,j}; e_i]\), where \(e_i\) is constructed from the hidden states of the last \(n\) tokens before the assistant token. We use \(n=2\).

We optimize probe parameters \(\phi\) using cross-entropy loss over the \(Q\) progress buckets. For an input representation \(\mathbf{h}_{i,j}\) (either \(h_{i,j}\) or \([h_{i,j}; e_i]\)) and one-hot target label \(y_{i,j} \in \{0,1\}^Q\), the loss is:

\begin{align}
    \mathcal{L}(\phi) = - \frac{1}{\sum_{i=0}^{B} T_i} \sum_{i=0}^{B} \sum_{j=0}^{T_i} \sum_{q=1}^{Q} y_{i, j, q} \times \\
    \log \left( \frac{\exp(f(\mathbf{h}_{i,j})_q)}{\sum_{k=1}^{Q} \exp(f(\mathbf{h}_{i, j})_k)} \right) \nonumber
\end{align}

\noindent where \(f(\mathbf{h}_{i,j})_q\) denotes the logit assigned to bucket \(q\) for token \(j\) in trace \(i\).

We restrict our analysis to linear probes to measure linearly accessible information in hidden states and thus we can avoid confounding representation quality with computation performed by a more expressive probe.

We select probe layers at representative early, middle, and late points of each model's transformer stack, matching roughly 29\%, 57\%, and 86\% depth across architectures for a fair comparison.

\begin{table*}[t]
\centering
\scriptsize
\resizebox{\linewidth}{!}{
\begin{tabular}{lcccccccccccc}
\toprule
\textbf{Probe}
& \multicolumn{4}{c}{\textbf{\deepscaler{}}}
& \multicolumn{4}{c}{\textbf{\qwenthreezeropointsixb{}}}
& \multicolumn{4}{c}{\textbf{\qwenthreefourbig{}}} \\
\cmidrule(lr){2-5} \cmidrule(lr){6-9} \cmidrule(lr){10-13}
& \textbf{Acc.} & \textbf{MAE} & \textbf{OOD Acc.} & \textbf{OOD MAE}
& \textbf{Acc.} & \textbf{MAE} & \textbf{OOD Acc.} & \textbf{OOD MAE}
& \textbf{Acc.} & \textbf{MAE} & \textbf{OOD Acc.} & \textbf{OOD MAE} \\
\midrule
Early $f_{\text{token}}$ & 0.285 & 1.745 & 0.129 & 2.998 & 0.281 & 1.538 & 0.202 & 1.923 & 0.320 & 1.318 & 0.189 & 2.267 \\
Early $f_{\text{q+token}}$ & 0.281 & 1.773 & 0.131 & 2.904 & 0.272 & 1.607 & 0.213 & 1.921 & 0.311 & 1.361 & 0.189 & 2.366 \\
Middle $f_{\text{token}}$ & \bestcell{0.307} & \bestcell{1.565} & 0.131 & 2.953 & \bestcell{0.301} & \bestcell{1.428} & \bestcell{0.252} & \bestcell{1.604} & \bestcell{0.359} & \bestcell{1.094} & \bestcell{0.212} & \bestcell{2.016} \\
Middle $f_{\text{q+token}}$ & 0.297 & 1.637 & \bestcell{0.136} & 2.835 & 0.293 & 1.471 & 0.240 & 1.705 & 0.348 & 1.147 & 0.206 & 2.099 \\
Late $f_{\text{token}}$ & 0.290 & 1.696 & 0.129 & 2.970 & 0.277 & 1.644 & 0.196 & 1.973 & 0.318 & 1.334 & 0.186 & 2.274 \\
Late $f_{\text{q+token}}$ & 0.283 & 1.702 & 0.132 & \bestcell{2.834} & 0.274 & 1.646 & 0.203 & 1.916 & 0.312 & 1.363 & 0.185 & 2.290 \\
\bottomrule
\end{tabular}
}
\caption{Linear probe performance across matched early, middle, and late normalized layer depths. MAE is computed from the probe's highest-probability progress bucket; OOD contains traces above 16K tokens. See Appendix~\ref{app:table-details} for probe feature definitions.}
\label{tab:probe-layer-results}
\end{table*}

\paragraph{Metrics}
We use top-1 accuracy to evaluate bucket classification performance and mean absolute error (MAE), following prior work \citep{huang2025leanprogressguidingsearchneural,ma2019selfmonitoringnavigationagentauxiliary,zhu2020visionlanguagenavigationselfsupervisedauxiliary}. MAE is computed as the absolute difference between the ground-truth progress bin and the bin assigned the highest probability.

For visualization, we additionally compute the expected progress value:
\[
\hat{p}_{i,j} = \sum_{q=1}^{Q} p(q \mid \mathbf{h}_{i,j}) \, c_q,
\]
where \(p(q \mid \mathbf{h}_{i,j})\) is the probability assigned to bucket \(q\), and \(c_q\) is the midpoint of bucket \(q\). This yields a continuous estimate of predicted progress from the probe's output distribution.

\paragraph{Results}

\begin{figure*}[t]
    \centering
    \includegraphics[width=\textwidth]{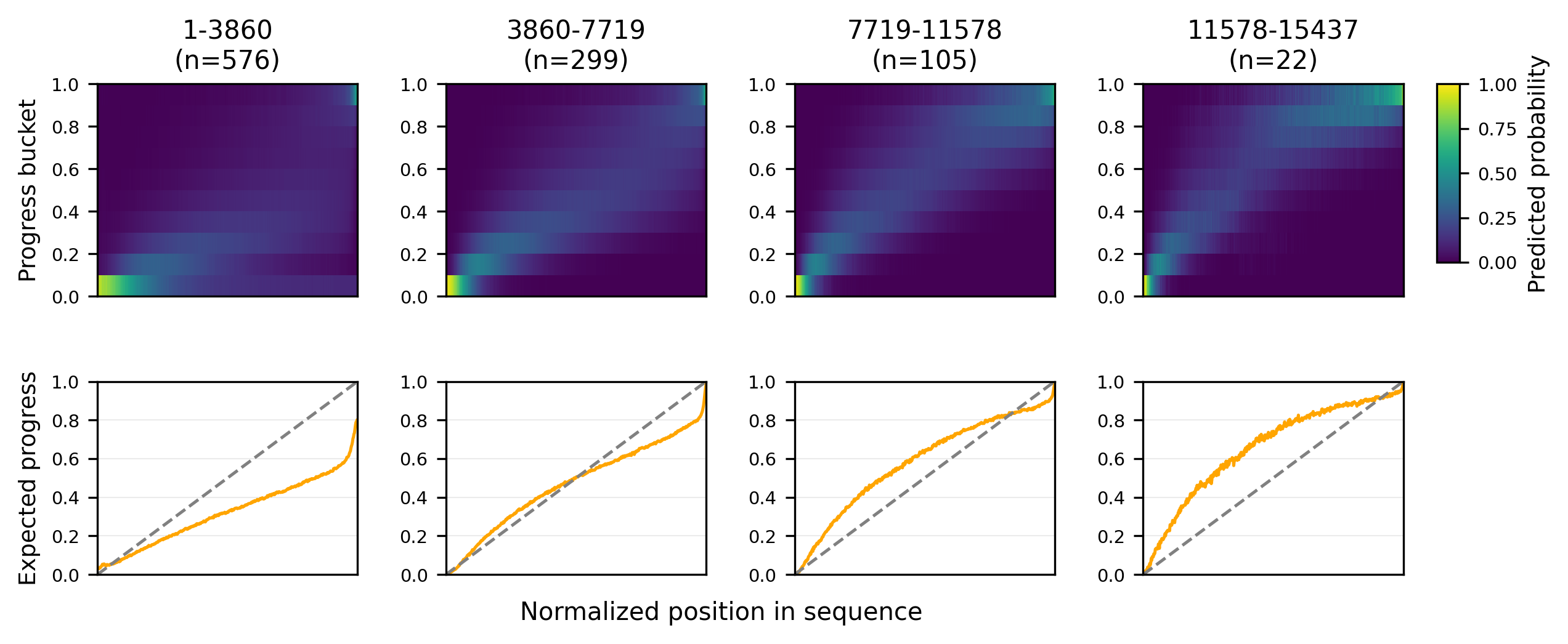}
    \caption{Probe predictions on the in-domain \deepscaler{} test split, grouped by reasoning-trace length. Top: average predicted probability over progress buckets. Bottom: expected progress curves, with the dashed line showing perfect calibration. See Appendix~\ref{app:probe-heatmap-aggregation} for how probabilities are computed and aggregated. The corresponding long-trace held-out visualization is shown in Appendix~\ref{img:heatmap_agg_proba_distb_ood}.}
    \label{img:heatmap_agg_proba_distb}
\end{figure*}

The top panels of Figure~\ref{img:heatmap_agg_proba_distb} show that the averaged predictive distributions place most probability mass near the diagonal, particularly near the beginning and end of reasoning traces, indicating lower uncertainty in those regions. The expected progress curves also closely track the ideal progress diagonal in the bottom panels. For a single-example visualization, we refer the reader to Appendix~\ref{img:proba_distb}.

Table~\ref{tab:probe-layer-results} shows that probe performance on \deepscaler{} is relatively consistent across layers, with accuracies around 29\%, substantially above the 10\% accuracy expected from random guessing over progress buckets. However, performance drops considerably on long-trace held-out examples, highlighting the difficulty of length generalization. The \qwenthreezeropointsixb{} probes show stronger OOD bucket accuracy, with the Middle layer token-only probe reaching 25.2\%. We observe the same length-generalization pattern for \qwenthreefourbig{}, but with stronger in-domain performance. Its best probe is the middle token-only probe, which reaches 35.9\% in-domain accuracy and 21.2\% OOD accuracy. Adding question embeddings does not consistently improve performance, suggesting that most progress-related information is already encoded in the hidden state at the current token position.

Overall, these results indicate that reasoning model hidden states encode information correlated with normalized trace position that is linearly recoverable, although this signal weakens on longer traces. This should be read as a representation diagnostic rather than conclusive evidence of semantic progress tracking, since the labels are derived from \(k/m\) and can partly reflect token-position and expected-length information.

\section{Incorporating Progress Estimation into the Reasoning Trace}

Our probing results suggest that reasoning model hidden states already encode information about reasoning progress. We next investigate whether explicit supervision can improve progress estimation by training models to report their progress directly within the reasoning trace.

\paragraph{Training Data}
We use OpenR1-Math \citep{openr1math2025}, where the reasoning traces were generated with DeepSeek-R1 \citep{deepseekai2025deepseekr1incentivizingreasoningcapability}. We augment each trace with explicit progress annotations inserted throughout the reasoning process. We construct these annotations by first splitting the reasoning into segments \(s_1, \ldots, s_n\) using the paragraph segmentation from \citet{frohmann-etal-2024-segment}. Then we insert a progress annotation of the form
\texttt{<progressbar>} \(a_i\) \texttt{</progressbar>}
after each segment, where \(a_i\) is the cumulative progress after segment \(s_i\):
\[
a_i =
\left\lfloor
100 \cdot
\frac{\sum_{j=1}^{i} |s_j|}
{\sum_{j=1}^{n} |s_j|}
\right\rfloor .
\]

In our main experiments, both the progress values \(a_i \in [0,100]\) and the \texttt{<progressbar>} delimiters are tokenized using the model's existing vocabulary rather than being added as new special tokens.

\paragraph{Evaluation Data}
We evaluate on MATH500 \citep{lightman2023lets}, AMC23 \citep{Amc23}, and a subset of OlympiadBench \citep{OlympiadBench}, which contain mathematical reasoning problems distinct from those seen during training. Together, these benchmarks span a broad range of problem difficulty and reasoning trace lengths, enabling us to study progress prediction across varying reasoning complexity.

To test whether progress supervision learned on mathematical traces transfers out of domain, we further evaluate on BIG-Bench Extra Hard (BBEH)~\citep{kazemi2025bigbenchextrahard}. We use the full 1{,}000-example core5 subset containing five non-arithmetic text-only tasks: zebra puzzles, web of lies, shuffled objects, spatial reasoning, and temporal sequence.

Further dataset details are provided in Appendix~\ref{app:dataset-splits}.

\paragraph{Metrics}
We evaluate both task accuracy and progress prediction error. Progress prediction is evaluated using MAE between predicted progress values and the true normalized position within the reasoning trace. For a completed reasoning trace \(\tau\), let \(\hat{p}_i \in [0,1]\) denote the \(i\)-th predicted progress value, emitted after prefix \(\mathbf{C}_{i}\). The corresponding realized progress is:
\[
g_i = \frac{|\mathbf{C}_{i}|}{|\tau|}.
\]

\noindent The MAE over the \(n\) predicted progress values is:
\[
\mathrm{MAE}
=
\frac{1}{n}
\sum_{i=1}^{n}
|\hat{p}_i - g_i| .
\]

\subsection{Models}

We fine-tune \deepscaler{}, \qwenthreezeropointsixb{}, and \qwenthreefourb{} using the two following variants.

\paragraph{Supervised Fine-Tuning (SFT)}
We fine-tune models using LoRA \citep{hu2021loralowrankadaptationlarge} applied to all linear layers, with \(r=256\) and \(\alpha=256\). Training uses the standard autoregressive cross-entropy loss, with tokens inside \texttt{<progressbar>}...\texttt{</progressbar>} annotations upweighted by a factor \(\gamma\). Specifically, let \(\tau = (x_1, \ldots, x_{|\tau|})\) denote a training sequence, and \(\mathcal{I}\) denote the set of token positions belonging to progress annotations, including both progress values and delimiter tokens. The loss is:
\begin{align}
L
&=
-\frac{1}{|\tau|}
\sum_{i=1}^{|\tau|}
w_i \log P_\theta(x_i \mid x_{<i})
\label{eq:sft-loss} \\
w_i
&=
1 + (\gamma - 1)\mathds{1}_{i \in \mathcal{I}} .
\nonumber
\end{align}

\noindent  In our experiments, we use \(\gamma = 5\) and a context length of \(2048\). Appendix~\ref{app:hyperparameters} lists the full fine-tuning hyperparameters in Table~\ref{tab:hyperparams}.

\paragraph{SFT with masking schedule}
\label{sec:sft-masking-schedule}
To reduce reliance on previously generated progress annotations, we additionally compare against models trained with a masking schedule. For each training sequence, each preceding progress annotation span, including all tokens within that span's \texttt{<progressbar>}...\texttt{</progressbar>} delimiters, is independently masked with probability \(\rho\). During training, \(\rho\) follows a cosine schedule from \(0\) to \(0.5\). This acts as a form of regularization that discourages the model from inferring progress purely from the spacing of prior progress markers. Without masking, the model could exploit the approximately regular spacing of progress annotations in the training data to estimate progress from token position alone rather than reasoning context. We analyze this effect further in Section~\ref{app:abl_masking_schedule}.

\paragraph{Baselines}
\label{sec:baselines}

For each emitted progress estimate, we compare the model against baselines that predict progress from token position and simple length statistics rather than reasoning content. \textit{Global mean length} divides the current prefix length by the mean completed trace length across the full evaluation set, then clips the result to the valid progress range. \textit{Task mean length} uses the same calculation, but replaces the global length estimate with the mean completed length for the corresponding model-family and task group. \textit{Task median length} is identical except that it uses the median completed length for that group, making it less sensitive to unusually long traces.

\textit{Previous marker} tests whether the model's own progress reports are self-consistent over time. For completed trace \(i\), let \(\mathbf{C}^{i}_{k_j}\) be the prefix where the \(j\)-th progress estimate is emitted, let \(\hat{p}_{i,j}\) be the clipped progress value emitted at that prefix, let \(G(i)\) be the corresponding model-family and task group, and let \(\tilde{m}_{G(i)}\) be that group's median completed length. The baseline is:
\[
\hat{p}^{\mathrm{prev}}_{i,j} =
\begin{cases}
c\left(|\mathbf{C}^{i}_{k_j}|\hat{p}_{i,j-1}/|\mathbf{C}^{i}_{k_{j-1}}|\right) & j>1,\\
c\left(|\mathbf{C}^{i}_{k_j}|/\tilde{m}_{G(i)}\right) & j=1,
\end{cases}
\]
where \(c\) clips values to \([0,1]\). Thus, after the first report, the baseline uses the previous report to infer the total length the model implicitly expected, then predicts the current progress from the current token position under that same implied length. For the first report, where no previous report exists, it falls back to the task median length baseline.

\subsection{Results}  

\begin{figure}[t]
    \centering
    \includegraphics[width=\columnwidth]{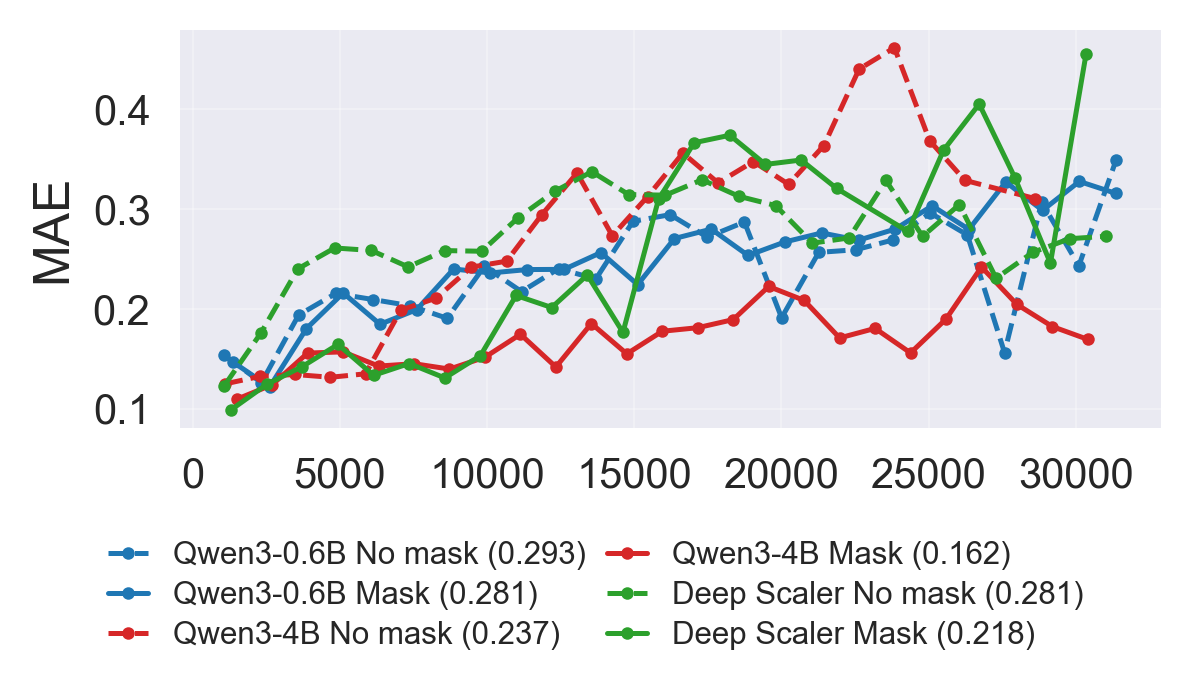}
    \caption{Prediction error (MAE) versus reasoning-trace length, binned into 25 groups across Qwen3-0.6B, Qwen3-4B, and Deep Scaler variants. Prediction error generally increases with completed trace length.}
    \label{fig:pred_by_seq}
\end{figure}

\paragraph{Masking improves progress prediction quality}
Table~\ref{tab:sft-special-tokens-masking} shows that the masking schedule usually improves progress prediction quality, although the effect varies by model family and evaluation domain. On the mathematical benchmarks, masking improves MAE for \qwenthreezeropointsixb{} and \qwenthreefourb{}, while the two \deepscaler{} progress-SFT variants are nearly tied. On BBEH, masking improves MAE for \deepscaler{} and \qwenthreefourb{} but not for \qwenthreezeropointsixb{}. At the same time, the unmasked \qwenthreefourb{} checkpoint achieves slightly higher downstream mathematical reasoning accuracy on MATH500, AMC23, and OlympiadBench, suggesting a trade-off between task performance and progress prediction quality. Overall, the masked \qwenthreefourb{} checkpoint gives the strongest progress prediction results, with MAE 0.161 on mathematical reasoning traces and 0.157 on BBEH.

\begin{table*}[t]
\centering
\scriptsize
\setlength{\tabcolsep}{2pt}
\begin{tabular}{>{\raggedright\arraybackslash}p{0.13\textwidth}>{\raggedright\arraybackslash}p{0.16\textwidth}rrrrr|rr}
\toprule
\textbf{Model}
& \textbf{Variant}
& \multicolumn{5}{c|}{\textbf{Math}} & \multicolumn{2}{c}{\textbf{BBEH}} \\
\cmidrule(lr){3-7} \cmidrule(lr){8-9}
 
&
& \textbf{MATH500 Acc.}
& \textbf{AMC23 Acc.}
& \textbf{OlympiadBench Acc.}
& \textbf{Macro Acc.}
& \textbf{MAE}
& \textbf{Acc.}
& \textbf{MAE} \\
\midrule

\multirow{4}{*}{\deepscaler{}}
& Original 
& 63.0 & 80.0 & 22.0 & 55.0 & N/A & 0.3 & N/A \\

& Dataset-only SFT 
& \textbf{63.0} & 62.5 & 15.3 & \textbf{46.9} & N/A & \textbf{0.1} & N/A \\

& Progress SFT 
& 56.8 & 60.0 & 14.0 & 43.6 & \textbf{0.294} & 0.0 & 0.265 \\

& Progress SFT + masking 
& 43.8 & \textbf{67.5} & \textbf{20.0} & 43.8 & 0.295 & 0.0 & \textbf{0.251} \\

\midrule

\multirow{4}{*}{\qwenthreezeropointsixb{}}
& Original 
& 59.0 & 45.0 & 16.0 & 40.0 & N/A & 3.3 & N/A \\

& Dataset-only SFT 
& 38.0 & \textbf{35.0} & 5.3 & 26.1 & N/A & \textbf{0.2} & N/A \\

& Progress SFT 
& 35.6 & 22.5 & 6.7 & 21.6 & 0.292 & \textbf{0.2} & \textbf{0.242} \\

& Progress SFT + masking 
& \textbf{41.6} & 32.5 & \textbf{8.0} & \textbf{27.4} & \textbf{0.280} & 0.0 & 0.256 \\

\midrule

\multirow{4}{*}{\qwenthreefourb{}}
& Original 
& 68.2 & 90.0 & 30.0 & 62.7 & N/A & 11.7 & N/A \\

& Dataset-only SFT 
& 70.6 & 77.5 & 24.7 & 57.6 & N/A & 0.0 & N/A \\

& Progress SFT 
& \textbf{70.8} & \textbf{82.5} & \textbf{27.3} & \textbf{60.2} & 0.231 & 0.0 & 0.193 \\

& Progress SFT + masking 
& 70.2 & 77.5 & 26.7 & 58.1 & \textbf{0.161} & \textbf{1.8} & \textbf{0.157} \\

\bottomrule
\end{tabular}
\caption{
Task accuracy and progress-prediction MAE results on mathematical reasoning benchmarks and BBEH core5. Bold indicates the best fine-tuned variant within each model family. See Appendix~\ref{app:table-details} for additional table details.
}
\label{tab:sft-special-tokens-masking}
\end{table*}
\paragraph{Accuracy degradation is largely explained by continued fine-tuning}
Table~\ref{tab:sft-special-tokens-masking} compares progress-reporting SFT against both the original models and continued SFT on the same dataset without progress annotations. This dataset-only SFT control isolates the effect of continued fine-tuning from the additional requirement of generating progress estimates. Across models, much of the downstream accuracy degradation on mathematical reasoning benchmarks is already present in the dataset-only SFT baseline, indicating that it cannot be attributed solely to progress prediction. Progress-token training can further affect reasoning performance, although the effect is not uniform across models or training settings. For example, masking improves mathematical reasoning accuracy for \qwenthreezeropointsixb{} relative to the corresponding unmasked progress-SFT variant. For \qwenthreefourb{}, the masked checkpoint improves progress MAE while slightly reducing macro accuracy relative to the unmasked run. Thus, the observed degradation should not be attributed solely to progress markers; part of it is already present in the dataset-only SFT control, and some may stem from the short 2048-token fine-tuning context.


\paragraph{Progress prediction generalizes beyond the training domain}
As shown in Table~\ref{tab:sft-special-tokens-masking}, task accuracy on BBEH remains near chance at the model scales we study. However, the primary purpose of evaluating on BBEH core5 is to test whether models trained on mathematical reasoning traces can still produce calibrated progress estimates on out-of-domain reasoning tasks. We observe that progress prediction remains measurable on BBEH, with MAE often comparable to, and sometimes lower than, mathematical progress MAE. This indicates that the progress-reporting behavior is not confined to the mathematical evaluation setting. Figure~\ref{fig:bbeh-progress-length} further shows per-marker BBEH prediction error grouped by completed trace length; unlike the mathematical-reasoning traces in Figure~\ref{fig:pred_by_seq}, BBEH does not show a consistent increase in error with length.

At the same time, Appendix~\ref{app:bbeh-position-baselines} shows that simple token-position and task-length baselines are highly competitive on BBEH, and often outperform the progress-reporting checkpoints. This suggests that BBEH traces contain strong structural regularities that make normalized trace position easier to estimate from length statistics alone.

We therefore interpret the BBEH results as evidence that progress reports remain parseable and numerically aligned out of domain, while leaving open how much of this alignment comes from content-sensitive progress estimation rather than trace-length regularities.


\paragraph{Neither Probes nor Fine-Tuned Models Uniformly Dominate}

To compare linear probes with fine-tuned progress-reporting models, we evaluate both on MATH500, AMC23, and OlympiadBench reasoning traces. Probe evaluation uses traces generated by the corresponding no-mask progress-SFT checkpoint. For each model family, we record the token positions where the SFT model emits progress estimates, remove all preceding \texttt{<progressbar>}...\texttt{</progressbar>} annotations from the prefix, and run the original pre-finetuning base model on this sanitized context to obtain hidden states. These hidden states are passed through the trained probe, and the probe's top-1 progress bin is converted into percentage-valued MAE. Thus, probes and no-mask SFT models are compared at identical token positions in identical generated traces, while probes are prevented from conditioning on earlier progress reports. Masked and unmasked progress-SFT models are otherwise evaluated on their own generated traces.

Table~\ref{tab:probe-vs-sft} shows a mixed picture. Fine-tuned progress prediction is not uniformly better than probe-based prediction, nor do probes uniformly dominate. For \deepscaler{}, the best in-domain learned result comes from a middle-layer probe with question and token features, while masked SFT gives the lowest OOD error among learned methods. For \qwenthreezeropointsixb{}, probes are strongest in both regimes. For \qwenthreefourb{}, masked SFT is strongest in-domain, but a middle-layer token-only probe is strongest OOD.

The position baselines are important context for interpreting these results. Mean-length baselines estimate progress from the current token position and an expected final trace length, while the previous-marker baseline extrapolates from the last emitted progress report. Their competitiveness, especially in-domain, suggests that much of the measured progress signal is tied to token position and expected trace length rather than necessarily reflecting content-sensitive estimates of remaining reasoning work. At the same time, the aggregate mathematical benchmark comparison in Appendix~\ref{app:math-position-baselines} shows that the strongest progress-SFT checkpoints can outperform these token-position baselines, indicating that direct progress supervision can add signal beyond length statistics in the mathematical setting.

\begin{table}[t]
\centering
\tiny
\setlength{\tabcolsep}{1pt}
\resizebox{\linewidth}{!}{
\begin{tabular}{lcc|cc|cc}
\toprule
\textbf{Probe or checkpoint}
& \multicolumn{2}{c}{\textbf{\deepscaler{}}}
& \multicolumn{2}{c}{\textbf{\qwenthreezeropointsixb{}}}
& \multicolumn{2}{c}{\textbf{\qwenthreefourb{}}} \\
\cmidrule(lr){2-3} \cmidrule(lr){4-5} \cmidrule(lr){6-7}
& \textbf{ID} & \textbf{OOD}
& \textbf{ID} & \textbf{OOD}
& \textbf{ID} & \textbf{OOD} \\
\midrule
Early $f_{\text{token}}$ & 15.93 & 39.83 & 26.53 & \textbf{15.00} & 19.80 & 21.62 \\
Early $f_{\text{q+token}}$ & 16.65 & 42.06 & 27.42 & 30.25 & 24.54 & 20.75 \\
Middle $f_{\text{token}}$ & 15.94 & 33.52 & 21.06 & 15.21 & 17.20 & \textbf{13.68} \\
Middle $f_{\text{q+token}}$ & \textbf{15.73} & 35.76 & \textbf{19.16} & 21.84 & 17.63 & 19.16 \\
Late $f_{\text{token}}$ & 17.67 & 32.01 & 22.28 & 18.62 & 20.99 & 15.50 \\
Late $f_{\text{q+token}}$ & 16.28 & 37.23 & 30.08 & 15.24 & 25.71 & 17.79 \\
\midrule
Global mean length & 20.28 & 35.98 & 24.06 & 23.91 & 20.59 & 42.95 \\
Dataset mean length & 18.53 & 31.71 & 23.69 & 22.19 & 18.77 & 36.74 \\
Dataset median length & 23.88 & 40.62 & 24.17 & 25.70 & 22.51 & 44.17 \\
Previous-marker extrap. & 27.57 & 31.34 & 22.68 & 32.04 & 19.63 & 39.08 \\
\midrule
SFT & 26.67 & 30.11 & 20.85 & 30.88 & 18.85 & 37.79 \\
SFT (mask) & 21.64 & \textbf{25.95} & 20.83 & 30.62 & \textbf{14.44} & 19.43 \\
\bottomrule
\end{tabular}
}
\caption{Linear-probe, token-position baseline, and SFT progress-prediction MAE on generated benchmark traces. ID contains traces below 16K tokens and OOD contains longer traces. See Appendix~\ref{app:table-details} for sample counts and baseline details.}
\label{tab:probe-vs-sft}
\end{table}

\section{Analysis}
\label{app:abl}

\subsection{Do Prior Progress Predictions Matter?}
\label{app:abl_masking_schedule}

We investigate whether conditioning on previously generated progress estimates helps refine future predictions or instead introduces compounding error over the course of a reasoning trace. Progress annotations introduce a possible shortcut: because markers are inserted at roughly regular token intervals, a model may be able to extrapolate from earlier progress reports and their positions in the sequence rather than infer progress from the reasoning content itself. This motivates the masking schedule introduced in Section~\ref{sec:sft-masking-schedule}.

We compare two inference settings on generated MATH500, OlympiadBench, and AMC23 traces: (i) \emph{Conditioned}, where the model can attend to its earlier progress predictions, and (ii) \emph{Unconditioned}, where all previous \texttt{<progressbar>}...\texttt{</progressbar>} annotations are removed before generating the next prediction. We evaluate both no-mask and masked progress-reporting checkpoints for \qwenthreezeropointsixb{}, \qwenthreefourb{}, and \deepscaler{} on 50 sampled traces per model.

\begin{table}
\centering
\scriptsize
\setlength{\tabcolsep}{2pt}
\resizebox{\linewidth}{!}{
\begin{tabular}{>{\raggedright\arraybackslash}p{0.30\columnwidth}>{\raggedright\arraybackslash}p{0.25\columnwidth}cc}
\toprule
\textbf{Model} & \textbf{Variant} & \textbf{Uncond.} & \textbf{Cond.} \\
\midrule
\multirow{2}{*}{\qwenthreezeropointsixb{}}
& No mask & 0.298 & 0.299 \\
& Mask & 0.264 & 0.266 \\
\midrule
\multirow{2}{*}{\deepscaler{}}
& No mask & 0.287 & 0.289 \\
& Mask & 0.229 & 0.318 \\
\midrule
\multirow{2}{*}{\qwenthreefourb{}}
& No mask & 0.264 & 0.239 \\
& Mask & 0.167 & 0.175 \\
\bottomrule
\end{tabular}
}
\caption{Conditioned versus unconditioned progress-prediction MAE for progress-reporting checkpoints. Lower is better. See Appendix~\ref{app:table-details} for sampling details.}\label{tab:conditional_vs_marginal}
\end{table}
Table~\ref{tab:conditional_vs_marginal} shows that conditioning on prior progress reports has no uniform effect across model families. For \qwenthreezeropointsixb{} and the no-mask \deepscaler{} checkpoint, conditioned and unconditioned MAE are nearly identical. The largest exception is masked \deepscaler{}, where the \emph{Unconditioned} setting achieves substantially lower MAE, suggesting that conditioning on prior progress estimates can accumulate prediction error over the course of a trace. For \qwenthreefourb{}, the masked checkpoint has much lower MAE than the no-mask checkpoint in both settings, although the no-mask checkpoint benefits more from using prior progress reports.

Together with the aggregate masked-versus-unmasked results in Table~\ref{tab:sft-special-tokens-masking}, these results suggest that masking is useful as a regularizer, while also showing that the value of conditioning on prior progress reports depends on the model family and checkpoint.

\begin{figure}[t]
    \centering
    \includegraphics[width=\columnwidth]{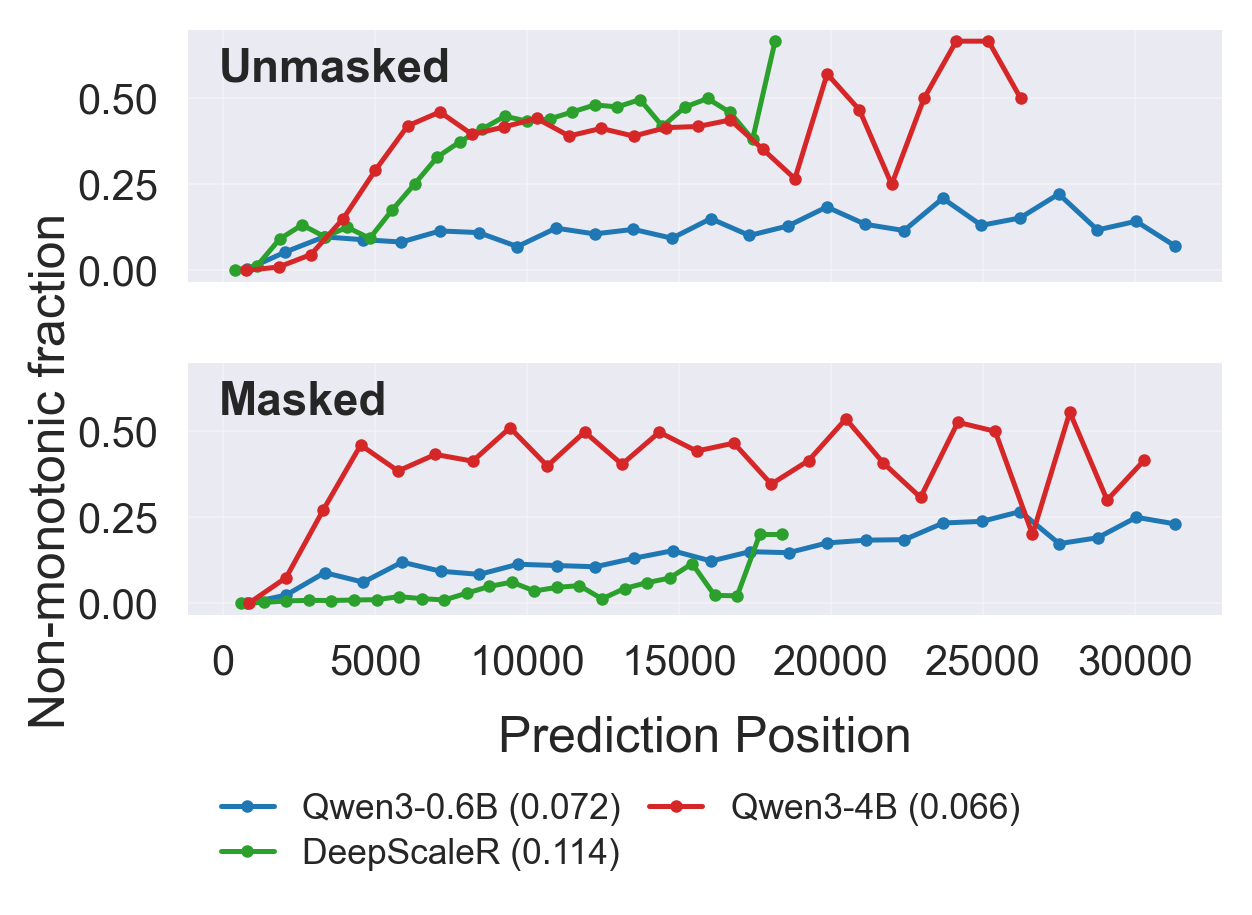}
    \caption{Fraction of non-monotonic predictions for different positions in the reasoning trace across Qwen3-0.6B, \deepscaler{}, and Qwen3-4B. The two panels compare unmasked training with the cosine masking schedule, with model families color-coded consistently with the sequence-length analysis.}
    \label{fig:monotonicity}
\end{figure}

Figure~\ref{fig:monotonicity} gives a complementary view of this behavior. If predictions were determined only by extrapolating from previous reports, they would be mostly monotonic. The observed non-monotonicity, and the differences between masked and unmasked checkpoints, suggest that models are not simply applying a fixed marker-spacing rule, although the effect varies by model family.
Appendix Figures~\ref{app:monotonicity_masked_schedule} and~\ref{app:monotonic_no_masking} show qualitative examples of these progress-prediction trajectories.

\subsection{How Predictable is Progress?}

Reasoning LM inference is stochastic, raising the question of how much irreducible uncertainty exists in progress estimation due to variation in possible continuation lengths. Given a question \(q\), we sample six partial reasoning prefixes \(\mathbf{C}_k\) from generated traces, where \(k\) denotes the prefix length. For each prefix, we then generate \(r=8\) sampled continuations
\(
\tau^i_\ell \sim P_\theta(\cdot \mid q, \mathbf{C}_k)
\)
using temperature 0.6, matching the recommended temperature for both DeepScaleR and Qwen3 reasoning inference \citep{deepscaler2025,qwen3}. We report this analysis on 500 MATH500 traces from each base model, yielding \(N=3000\) sampled prefixes and 24,000 continuations per model.
Appendix~\ref{app:mad_mapd_rollout_audit} and Table~\ref{tab:mad-mapd-rollout-audit} summarize the completed rollout jobs and validation checks.

For each sampled continuation \(\tau^i_\ell\), we compute the realized progress estimate \(g_{i,\ell}\), defined as the sampled prefix length divided by the completed rollout length. Let \(\bar{g}_i\) denote the average realized progress across the \(r\) sampled continuations from the same prefix.

To measure the intrinsic dispersion induced by continuation variability, we compute the Mean Absolute Deviation (MAD):
\begin{align}\label{eq:mad}
\text{MAD}
=
\frac{1}{r \cdot N}
\sum_{i=1}^{N}
\sum_{\ell=1}^{r}
\left|g_{i,\ell} - \bar{g}_i \right| .
\end{align}

We additionally compute the Mean Absolute Percentage Deviation (MAPD), which normalizes the same residuals by the average realized progress for each prefix:

\begin{align}\label{eq:mapd}
\text{MAPD}
=
\frac{1}{r \cdot N}
\sum_{i=1}^{N}
\sum_{\ell=1}^{r}
\frac{\left|g_{i,\ell} - \bar{g}_i \right|}{\bar{g}_i} .
\end{align}

\begin{figure}[t]
    \centering
    \includegraphics[width=\columnwidth]{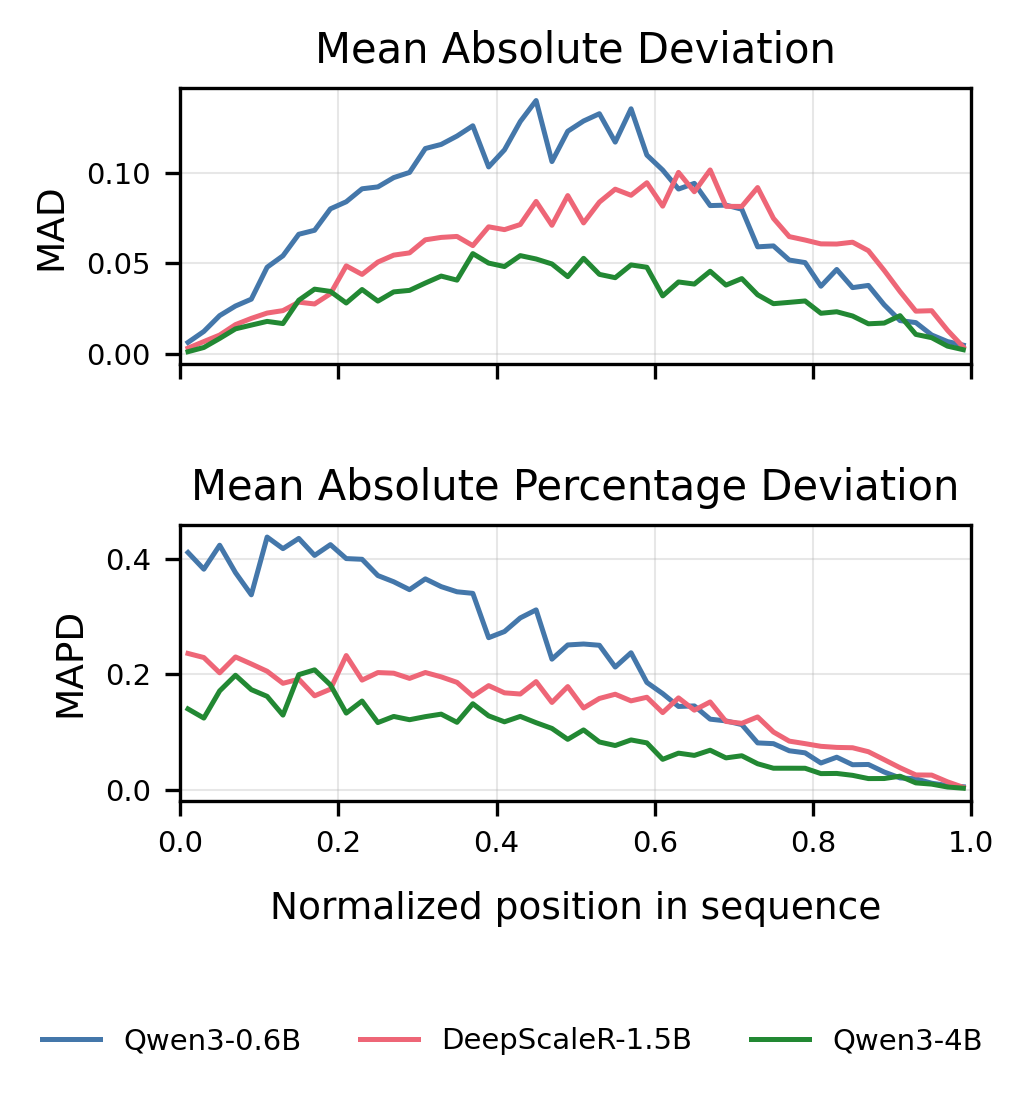}
    \caption{
       MAD (Equation~\ref{eq:mad}) and MAPD (Equation~\ref{eq:mapd}) for \qwenthreezeropointsixb{}, \deepscaler{}, and \qwenthreefourb{}, computed over 50 sequence length bins. Each panel overlays the three base models using separate colors.
    }
    \label{fig:trajectory-deviation}
\end{figure}

MAD measures the absolute spread in progress implied by different continuations from the same prefix, while MAPD measures the same variability relative to the average realized progress at that prefix. Figure~\ref{fig:trajectory-deviation} shows that rollout dispersion is model-dependent: \deepscaler{} and \qwenthreezeropointsixb{} have similar average MAD, while \qwenthreefourb{} has lower dispersion under the same sampling setup. MAPD generally decreases with normalized position because we are conditioning on more information, reducing uncertainty around the next token.

\begin{table}[t]
\centering
\scriptsize
\setlength{\tabcolsep}{2pt}
\resizebox{0.6\linewidth}{!}{
\begin{tabular}{lcc}
\toprule
\textbf{Model} & \textbf{MAD} & \textbf{MAPD} \\
\midrule
\qwenthreezeropointsixb{} & 0.065 & 0.168 \\
\deepscaler{} & 0.059 & 0.144 \\
\qwenthreefourb{} & 0.027 & 0.065 \\
\bottomrule
\end{tabular}
}
\caption{Rollout-dispersion summary for the predictability analysis. Each model is evaluated on 500 MATH500 reasoning traces, with six sampled prefixes per trace and eight sampled continuations per prefix.}
\label{tab:mad-mapd-in-progress}
\end{table}

MAD and MAPD should be read as estimates of sampled-continuation dispersion given the current prefix. If many plausible continuations have different final lengths, then the same prefix can correspond to different realized progress values, contributing to observed MAE. Comparing Table~\ref{tab:mad-mapd-in-progress} with Table~\ref{tab:sft-special-tokens-masking}, \qwenthreefourb{} has both the lowest rollout dispersion (MAD 0.027) and the lowest mathematical progress-prediction MAE (0.161), while \deepscaler{} and \qwenthreezeropointsixb{} have higher dispersion (MAD 0.059 and 0.065) and substantially higher best mathematical MAE (0.294 and 0.280). The remaining gap between MAD and observed MAE indicates room for better progress estimation beyond the uncertainty induced by stochastic continuation length.

\section{Conclusion and Future Research}
\label{sec:conclusion}

In this paper, we explore methods for predicting progress in reasoning models. Our experiments show that current models encode information correlated with normalized trace position: linear probes reach roughly 30\% accuracy over ten progress classes, and their errors are relatively localized. 

We introduce a simple strategy for directly incorporating progress prediction into reasoning LMs through SFT. The best progress-reporting checkpoint reaches 0.161 MAE on mathematical reasoning traces and outperforms token-position baselines in the aggregate mathematical benchmark comparison. However, the matched probe comparison is mixed: fine-tuned progress prediction does not uniformly dominate linear probes across model families and length regimes. On BBEH, progress reports remain parseable and numerically aligned, but strong token-position and task-length baselines show that out-of-domain MAE must be interpreted relative to structural controls. Sampled-continuation dispersion is lower than the observed MAE, suggesting that continuation-length variability explains part, but not all, of the remaining error.

\section*{Limitations}
\label{sec:limits}
Our study is limited to three compact models and mostly mathematical reasoning, with BBEH as the only non-mathematical evaluation setting. Prediction also degrades on traces longer than the training length distribution. Finally, progress MAE can reflect trace-length and marker-placement regularities; the BBEH baselines show that out-of-domain estimates should be interpreted relative to structural controls.

\bibliography{acl}

\clearpage
\appendix

\section{Additional Table Details}
\label{app:table-details}

For Table~\ref{tab:probe-layer-results}, early, middle, and late are matched normalized depths across the model families. The \(f_{\text{q+token}}\) probe concatenates the current-token hidden state with question embeddings, while \(f_{\text{token}}\) uses only the current-token hidden state.

\begin{figure*}[t]
    \centering
    \includegraphics[width=\textwidth]{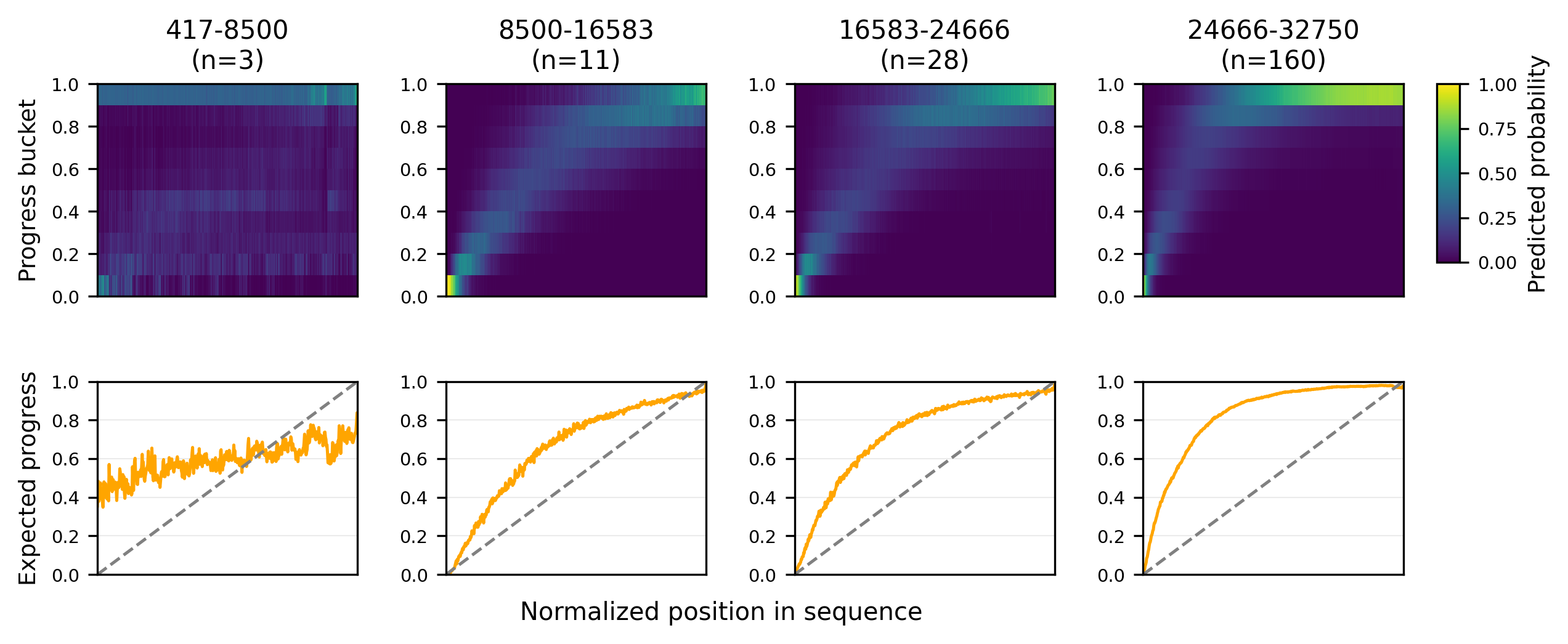}
    \caption{Probe predictions on long-trace held-out \deepscaler{} examples, restricted to reasoning traces longer than 16K tokens and grouped by reasoning-trace length. Top: average predicted probability over progress buckets. Bottom: expected progress curves, with the dashed line showing perfect calibration. See Appendix~\ref{app:probe-heatmap-aggregation} for how probabilities are computed and aggregated. This is the long-trace held-out counterpart to Figure~\ref{img:heatmap_agg_proba_distb}.}
    \label{img:heatmap_agg_proba_distb_ood}
\end{figure*}

\paragraph{Probe heatmap aggregation}
\label{app:probe-heatmap-aggregation}
Figures~\ref{img:heatmap_agg_proba_distb} and~\ref{img:heatmap_agg_proba_distb_ood} visualize the categorical distributions produced by the trained linear probe. At each token position \(j\) in trace \(i\), the probe logits \(f(\mathbf{h}_{i,j})\) are converted to progress-bucket probabilities with a softmax,
\[
p(q \mid \mathbf{h}_{i,j}) =
\frac{\exp(f(\mathbf{h}_{i,j})_q)}
{\sum_{k=1}^{Q} \exp(f(\mathbf{h}_{i,j})_k)}.
\]
We group traces into sequence-length buckets, with \(n\) denoting the number of examples in each bucket. Because traces have different lengths, each trace's per-token probability vectors are linearly interpolated onto a shared normalized position grid from 0 to 1. The heatmap value at a normalized position is then the arithmetic mean of these interpolated probability vectors across all traces in the length bucket. The bottom panels show the expected progress value computed from the same averaged distribution,
\[
\hat{p} = \sum_{q=1}^{Q} \bar{p}(q) c_q,
\]
where \(\bar{p}(q)\) is the averaged probability for bucket \(q\) at that normalized position and \(c_q\) is the midpoint of bucket \(q\). Figure~\ref{img:heatmap_agg_proba_distb} uses in-domain evaluation traces with at most 16K tokens, while Figure~\ref{img:heatmap_agg_proba_distb_ood} uses long-trace held-out examples with more than 16K tokens.

For Table~\ref{tab:sft-special-tokens-masking}, original models are included for reference only. Dataset-only SFT rows control for generic continued fine-tuning on the same data without progress markers. MAE is not applicable for original models and dataset-only SFT controls when they do not emit progress annotations.

For Table~\ref{tab:probe-vs-sft}, the probe rows use the same early, middle, and late normalized depths as Table~\ref{tab:probe-layer-results}. They use 439 ID and 36 OOD traces for \deepscaler{}, 388 ID and 30 OOD traces for \qwenthreezeropointsixb{}, and 630 ID and 40 OOD traces for \qwenthreefourb{}. The token-position baselines are computed on the no-mask generated traces at the same progress-report positions.

For Table~\ref{tab:conditional_vs_marginal}, conditioned scores use the progress reports generated in the original trace. Unconditioned scores resample each current progress report after removing earlier progress reports from the prefix. Results are computed over all progress predictions in 50 sampled traces per model, sampled from each model's generated MATH500, OlympiadBench, and AMC23 traces.

\section{Hyperparameters}\label{app:hyperparameters}
\begin{table}[H]
\centering
\scriptsize
\setlength{\tabcolsep}{2pt}
\begin{tabular*}{\columnwidth}{@{\extracolsep{\fill}}ll@{}}
\toprule
\textbf{Parameter} & \textbf{Value} \\
\midrule
learning rate & $1\times10^{-6}$ \\
lr scheduler & constant with warmup \\
warmup steps & 25 \\
num train epochs & 1 \\
LoRA rank & 256 \\
LoRA alpha & 256 \\
progress-token loss weight \(\gamma\) & 5 \\
context length & 2048 \\
num generations & 4 \\
temperature & 1.0 \\
max steps & 500 \\
bf16 & True \\
max grad norm & 0.1 \\
max completion length & 8096 \\
$\delta$ & 4 \\
\bottomrule
\end{tabular*}
\caption{Hyperparameters used in the fine-tuning experiments.}
\label{tab:hyperparams}
\end{table}

\section{Dataset Splits}\label{app:dataset-splits}

The model-specific trace datasets are generated from the same 25,175 OpenR1-Math questions, split into 22,657 training questions and 2,518 evaluation questions. For each model, we generate one reasoning trace per question. We then split each train/evaluation subset by trace length: traces with at most 16K tokens are treated as in-domain, while traces longer than 16K tokens are treated as long-trace held-out examples. The question split is therefore identical across models; only the short/long counts vary because the models produce different trace lengths for the same questions. Table~\ref{tab:dataset-split-sizes} gives the resulting counts.

\begin{table*}[t]
\centering
\scriptsize
\setlength{\tabcolsep}{2pt}
\begin{adjustbox}{max width=\textwidth}
\begin{tabular}{lrrrrrrr}
\toprule
\textbf{Model} & \multicolumn{1}{c}{\textbf{All Questions}} & \multicolumn{3}{c}{\textbf{Train Questions}} & \multicolumn{3}{c}{\textbf{Evaluation Questions}} \\
\cmidrule(lr){2-2}\cmidrule(lr){3-5}\cmidrule(lr){6-8}
& \textbf{Total} & \textbf{Total} & \textbf{Short} & \textbf{Long} & \textbf{Total} & \textbf{Short} & \textbf{Long} \\
& & & \textbf{($\leq$16K)} & \textbf{($>$16K)} & & \textbf{($\leq$16K)} & \textbf{($>$16K)} \\
\midrule
\deepscaler{} & 25,175 & 22,657 & 21,519 & 1,138 & 2,518 & 2,385 & 133 \\
\qwenthreezeropointsixb{} & 25,175 & 22,657 & 22,397 & 260 & 2,518 & 2,476 & 42 \\
\qwenthreefourbig{} & 25,175 & 22,657 & 17,013 & 5,644 & 2,518 & 1,878 & 640 \\
\bottomrule
\end{tabular}
\end{adjustbox}
\caption{Dataset split accounting for the OpenR1-Math-derived trace datasets used in the linear probe experiments. Train and evaluation question counts are fixed across models. Within each split, traces with at most 16K tokens are used for in-domain training or evaluation, while traces over 16K tokens form the long-trace held-out evaluation.}
\label{tab:dataset-split-sizes}
\end{table*}

\section{Math Token-Position Baselines}
\label{app:math-position-baselines}

Table~\ref{tab:math-position-baselines} reports the token-position baselines defined in Section~\ref{definition} on the mathematical benchmark traces used in Table~\ref{tab:sft-special-tokens-masking}.

\begin{table*}[t]
\centering
\scriptsize
\setlength{\tabcolsep}{2pt}
\begin{adjustbox}{max width=\textwidth}
\begin{tabular}{llrrrrr}
\toprule
\textbf{Model} & \textbf{Variant} & \textbf{Progress Model} & \textbf{Global} & \textbf{Dataset Mean} & \textbf{Dataset Median} & \textbf{Prev. Marker} \\
\midrule
\multirow{2}{*}{\qwenthreezeropointsixb{}}
& No masking & 0.293 & 0.239 & 0.224 & 0.255 & 0.305 \\
& Masking & 0.281 & 0.243 & 0.226 & 0.275 & 0.291 \\
\midrule
\multirow{2}{*}{\deepscaler{}}
& No masking & 0.281 & 0.270 & 0.241 & 0.310 & 0.292 \\
& Masking & 0.218 & 0.264 & 0.238 & 0.286 & 0.220 \\
\midrule
\multirow{2}{*}{\qwenthreefourb{}}
& No masking & 0.237 & 0.263 & 0.234 & 0.280 & 0.246 \\
& Masking & 0.162 & 0.270 & 0.238 & 0.275 & 0.168 \\
\bottomrule
\end{tabular}
\end{adjustbox}
\caption{Aggregate mathematical benchmark progress MAE compared with token-position baselines on MATH500, AMC23, and OlympiadBench traces. Lower is better.}
\label{tab:math-position-baselines}
\end{table*}

\section{BBEH Token-Position Baselines}
\label{app:bbeh-position-baselines}

The table below compares BBEH progress-reporting models against the token-position baselines defined in Section~\ref{definition}.

These baselines are diagnostic rather than deployable online predictors because they use realized corpus-level information. They show that, on BBEH, much of the measured progress signal is captured by simple length regularities. The progress-reporting models do not consistently outperform these token-position controls, which suggests that improving out-of-domain progress prediction requires reducing reliance on completed-length priors.

\begin{table*}[t]
\centering
\scriptsize
\setlength{\tabcolsep}{2pt}
\begin{adjustbox}{max width=\textwidth}
\begin{tabular}{llrrrrr}
\toprule
\textbf{Model} & \textbf{Variant} & \textbf{Progress Model} & \textbf{Global} & \textbf{Task Mean} & \textbf{Task Median} & \textbf{Prev. Marker} \\
\midrule
\multirow{2}{*}{\qwenthreezeropointsixb{}}
& No masking & 0.242 & 0.089 & 0.041 & 0.008 & 0.248 \\
& Masking & 0.256 & 0.086 & 0.031 & 0.009 & 0.260 \\
\midrule
\multirow{2}{*}{\deepscaler{}}
& No masking & 0.265 & 0.130 & 0.127 & 0.120 & 0.269 \\
& Masking & 0.251 & 0.088 & 0.064 & 0.023 & 0.257 \\
\midrule
\multirow{2}{*}{\qwenthreefourb{}}
& No masking & 0.193 & 0.084 & 0.089 & 0.080 & 0.212 \\
& Masking & 0.157 & 0.091 & 0.085 & 0.061 & 0.157 \\
\bottomrule
\end{tabular}
\end{adjustbox}
\caption{BBEH progress-prediction MAE compared with diagnostic token-position baselines. Lower is better. Global uses the global mean completed length; Task Mean and Task Median use per-model-family, per-task completed lengths; Prev. Marker extrapolates from the previous generated progress report.}
\label{tab:bbeh-position-baselines}
\end{table*}

\clearpage

\begin{figure*}
    \centering
    \includegraphics[width=\textwidth]{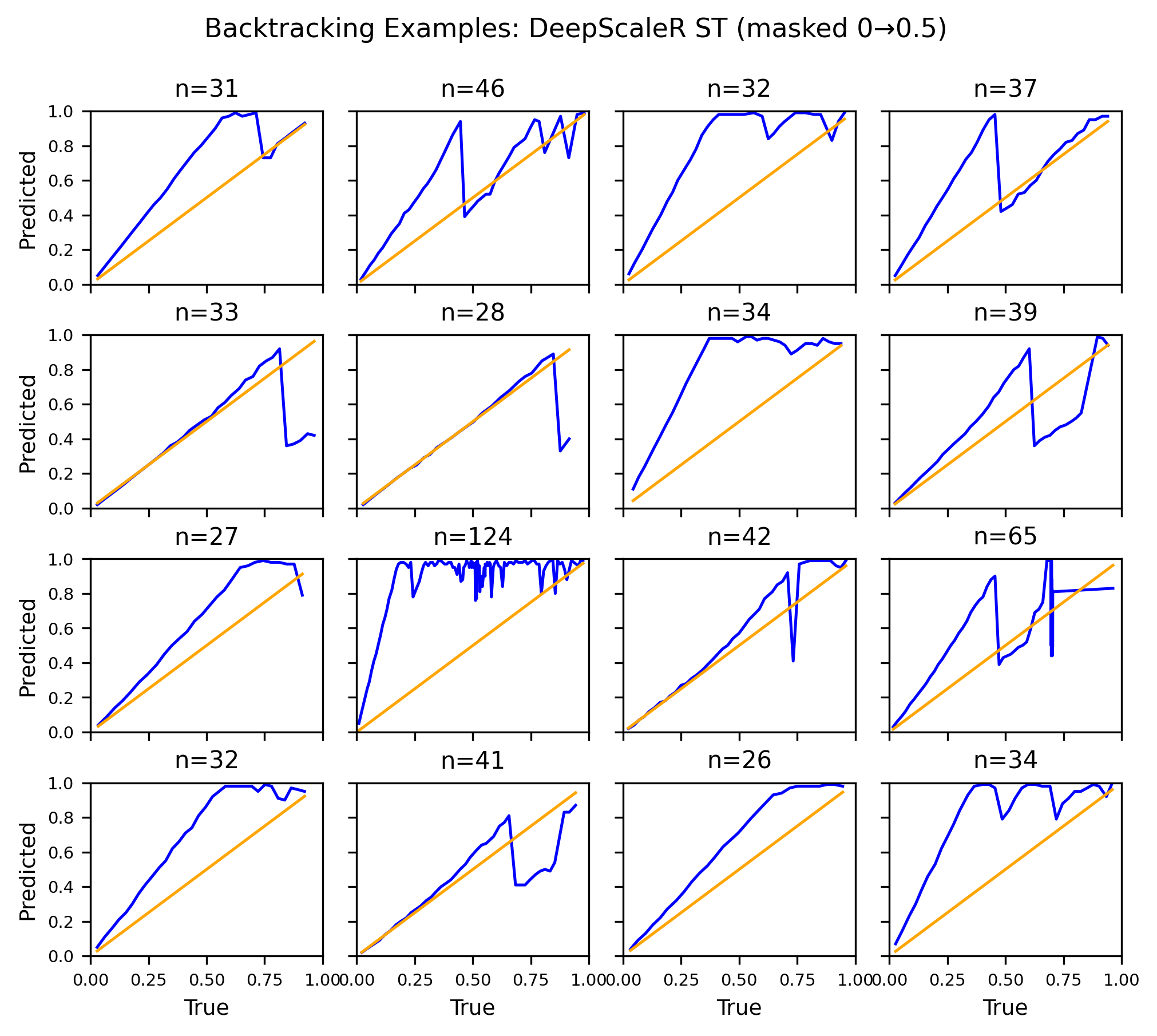}
    \caption{Sixteen examples of backtracking. The line with slope 1 indicates perfect prediction for the model trained with progress annotations and a mask.}
    \label{app:monotonicity_masked_schedule}
\end{figure*}

\begin{figure*}
    \centering
    \includegraphics[width=\textwidth]{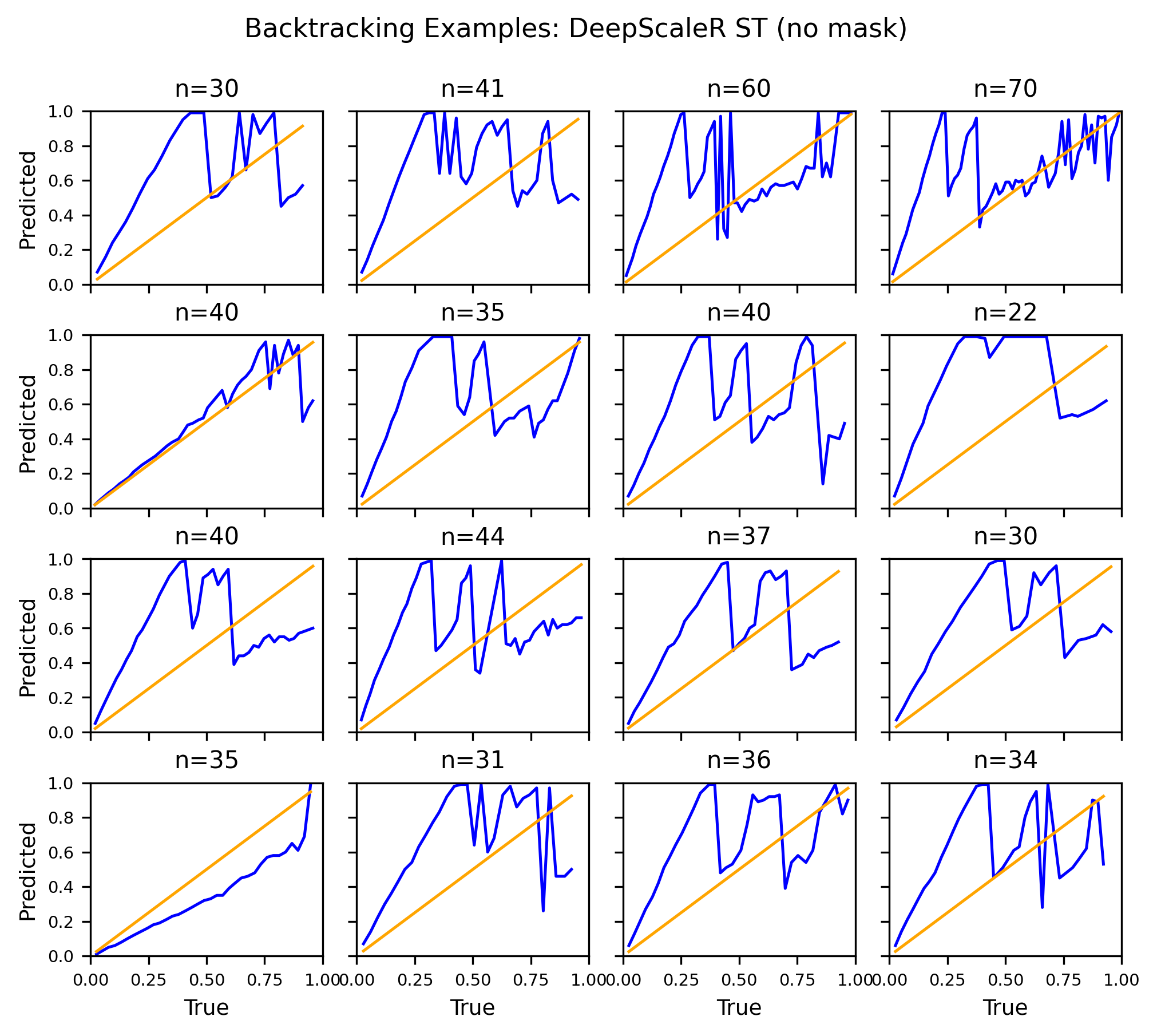}
    \caption{Example progress-prediction trajectories for the unmasked progress-reporting model.}
    \label{app:monotonic_no_masking}
\end{figure*}

\begin{figure*}
    \centering
    \includegraphics[width=\textwidth]{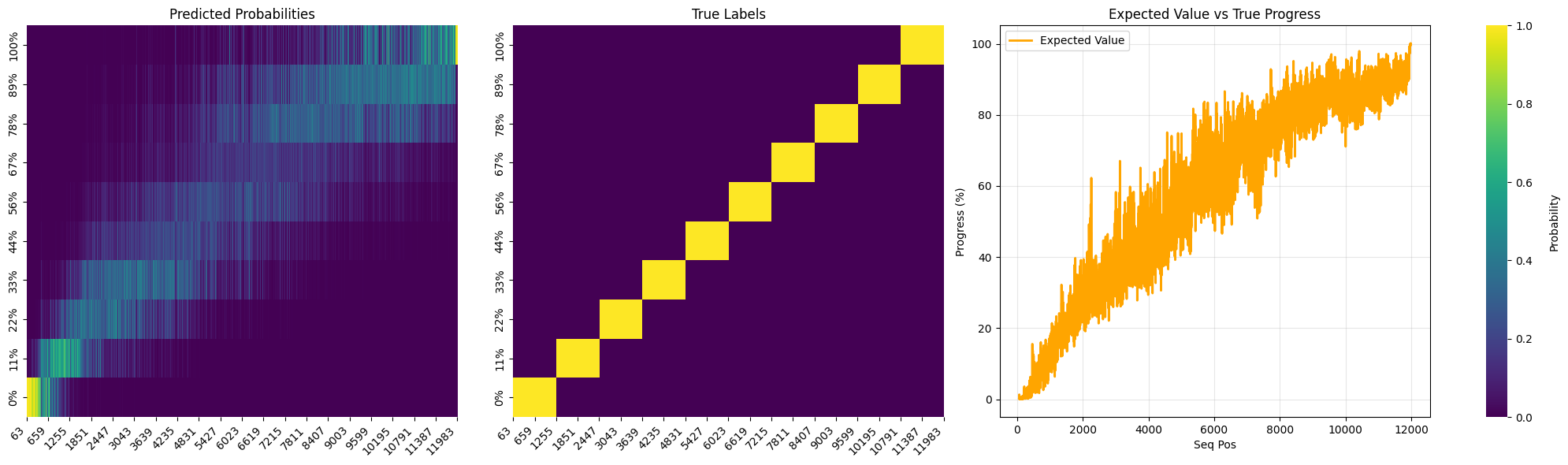}
    \caption{Single-trace linear probe visualization for a 12K-token sequence. Left: predicted progress-bucket probabilities. Center: ground-truth progress labels. Right: expected progress from the probe distribution.}
    \label{img:proba_distb}
\end{figure*}

\begin{figure*}[t]
    \centering
    \includegraphics[width=\textwidth]{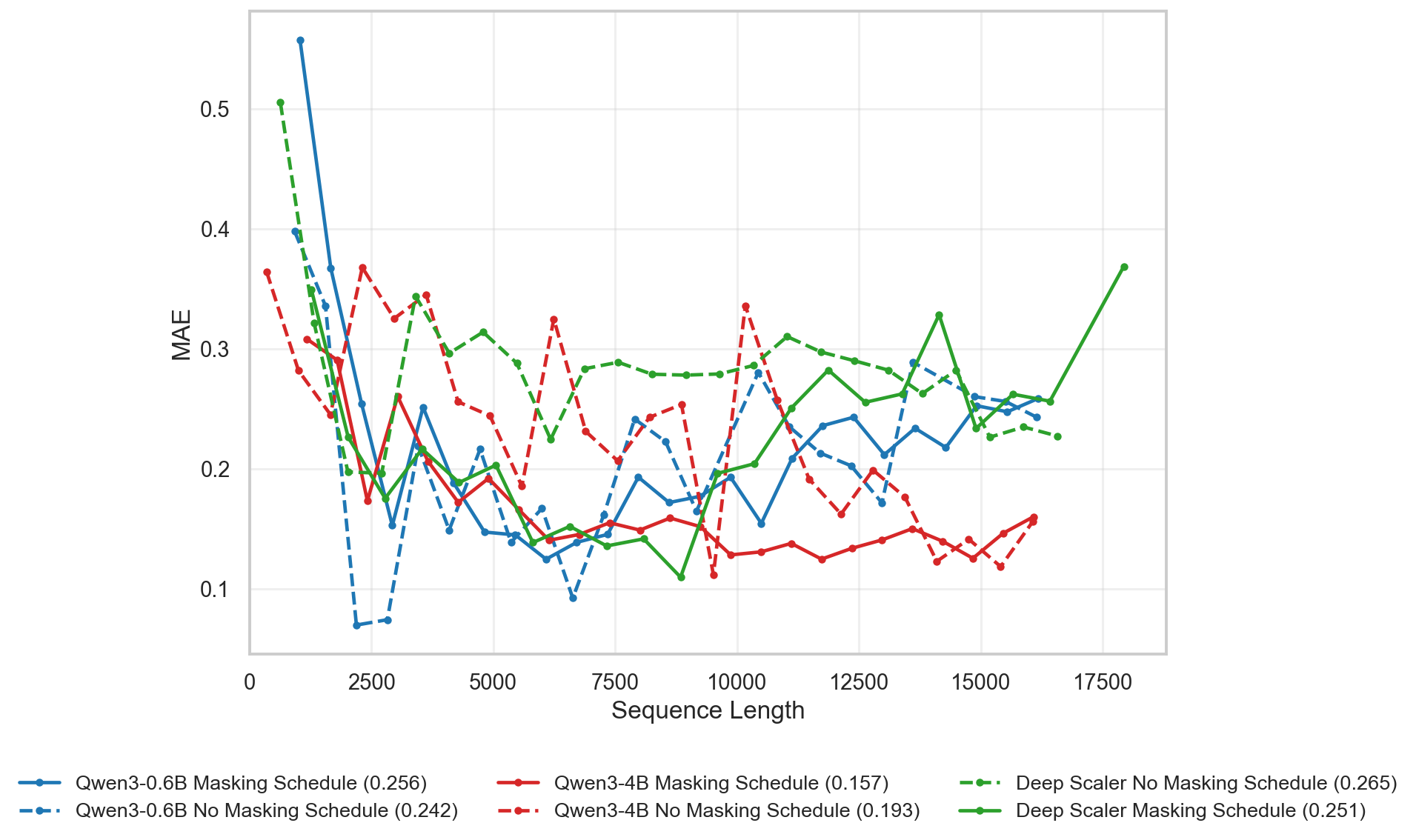}
    \caption{BBEH prediction error (MAE) versus sequence length, binned into 25 groups, across Qwen3-0.6B, Qwen3-4B, and Deep Scaler variants. Each point is the mean absolute error over progress markers from traces in the corresponding length bin.}
    \label{fig:bbeh-progress-length}
\end{figure*}

\clearpage

\section{Analyzing Dispersion}
\label{app:mad_mapd_rollout_audit}

For the rollout-dispersion analysis in Section~\ref{tab:mad-mapd-in-progress}, we used the base-model reasoning traces for each evaluated model and sampled six prefix positions per trace. Each prefix was branched into eight continuations at temperature 0.6. The completed runs used 500 traces per model, yielding 3000 sampled prefixes and 24,000 rollouts for each model.

\begin{table}[H]
\centering
\scriptsize
\setlength{\tabcolsep}{2pt}
\begin{tabular*}{\columnwidth}{@{\extracolsep{\fill}}lrrrr@{}}
\toprule
\textbf{Model} & \textbf{Job} & \textbf{Runtime} & \textbf{Prefixes} & \textbf{Rollouts} \\
\midrule
\qwenthreezeropointsixb{} & 5248 & 09:46:49 & 3000 & 24000 \\
\deepscaler{} & 2921 & 03:23:31 & 3000 & 24000 \\
\qwenthreefourb{} & 5249 & 15:46:49 & 3000 & 24000 \\
\bottomrule
\end{tabular*}
\caption{Audit summary for the sampled-continuation runs used in the MAD/MAPD analysis. The Qwen jobs are the corrected async vLLM runs, after verifying that every sampled prefix saved all eight rollouts.}
\label{tab:mad-mapd-rollout-audit}
\end{table}

We validated that each completed JSON contains six sampled prefixes and that every prefix contains exactly eight completed continuations. The estimate JSON files are tracked with Git LFS.

\section{Use of large language models}
\label{app:llm-use}

This research was conducted with assistance from large language models across three primary areas: implementation, ideation, and research. For implementation, we used large language models to help with coding tasks, debugging, and text polishing, including identifying spelling errors and improving clarity. For ideation, the models provided suggestions for potentially interesting experiments and research directions. For research, we employed OpenAI's Deep Research feature to gather information about related papers and work in the field. Additionally, large language models assisted with data visualization and plotting tasks throughout the project.

\section{Model Size And Budget}
\label{app:model-size-budget}

Our subject models are \qwenthreezeropointsixb{} with $0.6\text{B}$ parameters, \deepscaler{} with $1.5\text{B}$ parameters, and \qwenthreefourb{} with $4\text{B}$ parameters. The experiments in this paper use LoRA fine-tuning and sampled inference runs on A100-80GB GPUs. We do not use compute budget as an experimental variable.

\end{document}